\documentclass{article}

\usepackage{arxiv}
\usepackage[utf8]{inputenc} 
\usepackage[T1]{fontenc}    
\usepackage{hyperref}       
\usepackage{url}            
\usepackage{booktabs}       
\usepackage{amsfonts}       
\usepackage{nicefrac}       
\usepackage{microtype}      
\usepackage{lipsum}
\usepackage{graphicx}

\usepackage{xcolor}         
\usepackage{amsmath} 
\usepackage[linesnumbered,ruled,vlined]{algorithm2e}
\usepackage{booktabs}
\usepackage{multirow}
\usepackage{array}
\usepackage{xcolor}
\usepackage{colortbl} 
\usepackage{graphicx}
\usepackage[
  style=authoryear, 
  backend=biber,
  maxnames=1,       
  minnames=1,       
  maxbibnames=99,   
]{biblatex}
\DeclareNameFormat{author}{%
  \ifcase\value{listcount}
    \usebibmacro{name:family}
      {\namepartfamily}
      {\namepartgiven}
      {\namepartprefix}
      {\namepartsuffix}%
    \usebibmacro{name:andothers}
  \fi
}

\DeclareCiteCommand{\cite}
  {\usebibmacro{prenote}}
  {\usebibmacro{citeindex}%
   \printnames{author}%
   \setunit{\nameyeardelim}%
   \printfield{year}}
  {\multicitedelim}
  {\usebibmacro{postnote}}

\DeclareCiteCommand{\textcite}
  {\boolfalse{cbx:parens}}
  {\usebibmacro{citeindex}%
   \printnames{author}%
   \setunit{\nameyeardelim}%
   \printfield{year}}
  {\ifbool{cbx:parens}
     {\bibcloseparen\global\boolfalse{cbx:parens}}
     {}%
   \multicitedelim}
  {\usebibmacro{textcite:postnote}}

\definecolor{myblue}{rgb}{0.816, 0.902, 0.965}
\date{}  

\title{From Language to Logic: A Bi-Level Framework for
Structured Reasoning}

\author{
Keying Yang \\
  Tongji University\\
  \texttt{yky@tongji.edu.cn}
   \And
Hao Wang \\
  Tongji University\\
  \texttt{2432196@tongji.edu.cn}
  \And
Kai Yang \\
  Tongji University\\
\texttt{kaiyang@tongji.edu.cn}
}

\begin{document}
\maketitle

\vspace{-5pt}

\begin{abstract}
Structured reasoning over natural language inputs remains a core challenge in artificial intelligence, as it requires bridging the gap between unstructured linguistic expressions and formal logical representations. In this paper, we propose a novel \textbf{bi-level framework} that maps language to logic through a two-stage process: high-level task abstraction and low-level logic generation. At the upper level, a large language model (LLM) parses natural language queries into intermediate structured representations specifying the problem type, objectives, decision variables, and symbolic constraints. At the lower level, the LLM uses these representations to generate symbolic workflows or executable reasoning programs for accurate and interpretable decision making. The framework supports modular reasoning, enforces explicit constraints, and generalizes across domains such as mathematical problem solving, question answering, and logical inference. We further optimize the framework with an end-to-end {bi-level} optimization approach that jointly refines both the high-level abstraction and low-level logic generation stages. Experiments on multiple realistic reasoning benchmarks demonstrate that our approach significantly outperforms existing baselines in accuracy, with accuracy gains reaching as high as 40\%. Moreover, the bi-level design enhances transparency and error traceability, offering a promising step toward trustworthy and systematic reasoning with LLMs.
\end{abstract}

\vspace{-5pt}

\section{Introduction}
Structured reasoning is a critical challenge in large language models (LLMs), aiming to guide models to solve problems or derive conclusions through a systematic, step-by-step process, rather than relying on intuition or pattern matching. Early advances like the Chain-of-Thought (CoT) paradigm \parencite{wei2022chain} have gained attention by breaking complex problems into explicit intermediate steps. Subsequently, several efforts, such as TOT \parencite{yao2023tree} and GOT \parencite{besta2024graph} search, planning \parencite{hao2023RAP}, task decomposition \parencite{zhou2022least}, and meta-thinking methods (e.g., Meta‑CoT \parencite{xiang2025towards}), have been proposed to help organize reasoning trajectories for longer CoT. Recent advanced reasoning models such as deepseek-R1 \parencite{guo2025deepseek} and OpenAI-o1 \parencite{jaech2024openai} leverage increased inference time and extended CoT to enhance reasoning accuracy. These models are characterized by their ability to emulate human reasoning through a slower, more deliberate thought process. However, they still depend on unstructured natural‑language steps and lack an explicit mechanism for capturing a problem’s underlying logic. Therefore, despite significant performance improvements, these approaches have exposed critical challenges: inefficiency caused by overthinking \parencite{chen2024not}, heavy dependence on long contexts \parencite{yang2025pencil}, and reduced structure and interpretability as reasoning chains lengthen.

Leveraging code’s computational and logical strengths, although prior work has explored integrating LLMs with symbolic solver or executable code to handle complex computations \parencite{chen2022program, gao2023pal,xiao2024chain} or assist in solving specific logical problems \parencite{wu2022autoformalization, ye2023satlm, pan2023logic}, these methods are primarily restricted to logical reasoning and mathematical tasks. This limitation arises from their inability to disentangle implicit reasoning signals from noisy, ambiguous information inherent in abstract, complex problems. For instance, methods relying on fine-tuning or prompting for interpreter-based answering \parencite{gou2024tora, lu2024mathcoder2, chen2022program, gao2023pal} struggle to generalize, as they are tailored to specific task types and lack the flexibility to adapt to broader contexts. Similarly, approaches that combine large language models (LLMs) with logic programming (LP) languages face additional hurdles: they are encumbered by the unique syntax and interaction protocols of each language environment, severely limiting their generality and practical applicability.

Unlike the constrained approaches of prior methods, human experts, by contrast, tackle complex problems not just by prolonging deliberation but by deeply understanding their logical structure. They abstract and modeling problems—often using formal logical or mathematical frameworks to define variables, constraints, and objectives—before applying analytical or computational methods to solve them. This “modeling and solving” paradigm, prevalent in fields like mathematics \parencite{bender2000introduction} and operations research \parencite{churchman1957introduction}, inspires our approach. Drawing from this, we propose \textit{Lang2Logic}, a novel bilevel structured reasoning framework. Lang2Logic decomposes reasoning into two layers: a high-level \textit{Optimization-Guided Formalization(OGF) LLM}, which transforms natural-language queries into structured formal models (specifying problem type, variables, constraints, and objectives). In the lower layer, the \textit{Logic Generation(LG) LLM} first constructs logical representations, such as rule-based workflows or constraint-driven solution paths, and then translates these into executable Python code, which serves as a universal symbolic workflow logic generation. This design enables modular reasoning across diverse domains, while ensuring the reasoning process is both systematic and transparent.

To further enhance generalization and collaboration, we design a two‑stage training strategy: first, we cold‑start the OGF LLM via supervised fine‑tuning on a model‑augmented dataset; then, we propose a bilevel optimization algorithm to jointly train our framework. Extensive experiments on 9 challenging reasoning benchmarks spanning causal reasoning, logical puzzles, spatial reasoning, temporal reasoning, and mathematical reasoning with multiple backbone LLMs show that Lang2Logic achieves on average over 10\% absolute accuracy gains compared to the strongest baselines, with particularly large improvements on more complex problems. Here, we summarize our contributions as follows:
\begin{itemize}
\item We introduce a novel two-stage architecture that decomposes structured reasoning into high-level task abstraction and low-level logic generation, enabling systematic translation from natural language to executable symbolic representations.
\item The proposed framework is jointly optimized through an end-to-end bi-level optimization algorithm, allowing coordinated learning across abstraction and execution layers, and leading to improved reasoning accuracy and consistency.
\item Extensive experiments on critical reasoning benchmarks demonstrate that our framework significantly improves accuracy while producing clearer, more interpretable reasoning processes.
\end{itemize}

\begin{figure}[ht]
    \centering
    \includegraphics[width=\textwidth]{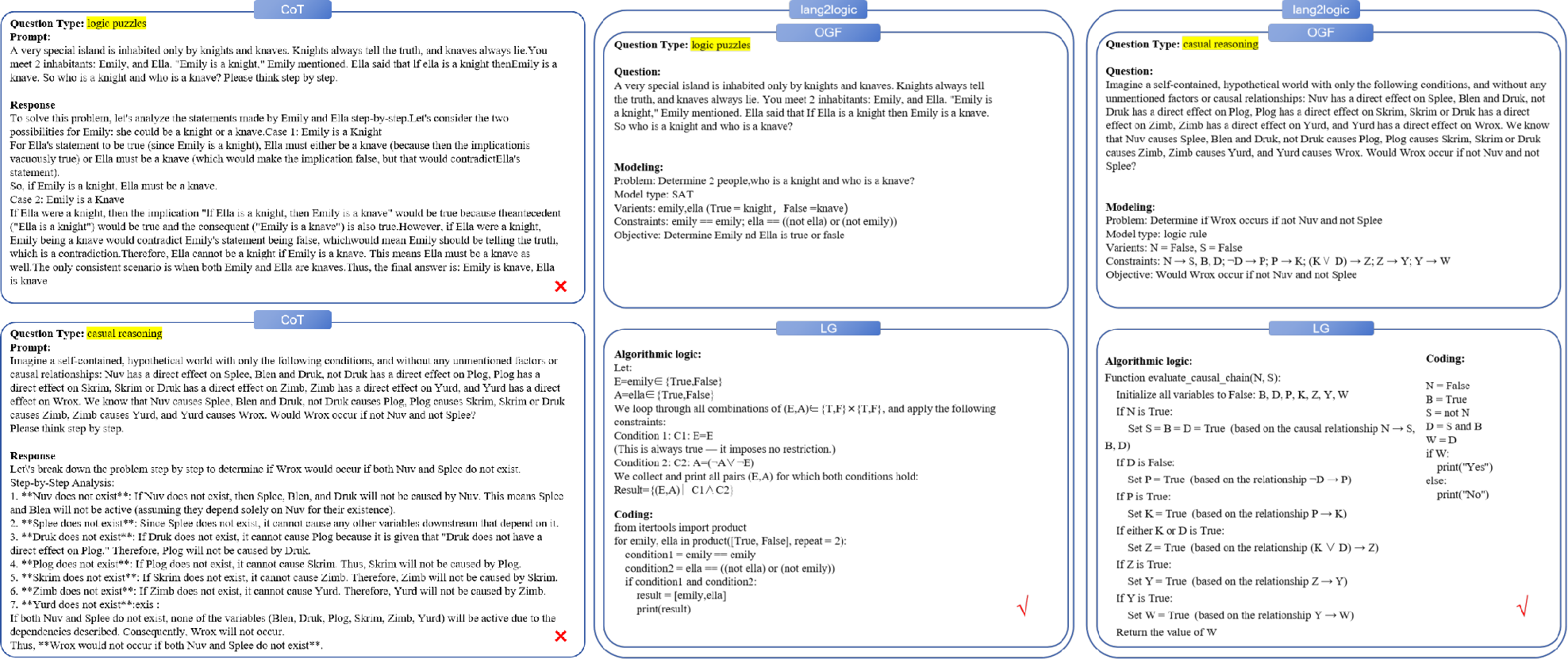} 
    \caption{Two examples for casual reasoning tasks and logic puzzles.The reasoning process of Chain-of-Thought (CoT) is lengthy and difficult to read, often failing to correctly understand the problem's underlying inference rules. The Lang2Logic framework provides a clear logical reasoning process through task abstraction and logic generation.}
    \label{fig:example}
\end{figure}
\vspace{-10pt}

\section{Related Work}
\label{gen_inst}
\paragraph{Structured Reasoning Paradigm}  
A key distinction of structured reasoning approaches stems from how reasoning steps are managed during generation. Starting from the Chain-of-Thought (CoT) paradigm \parencite{wei2022chain}, which proposes to elicit LLM thinking step by step, different reasoning paradigms have emerged to bolster reasoning performance, mainly including: planning \parencite{hao2023RAP,wang2023plan}, problem decomposition \parencite{zhou2022least,khot2022decomposed}, analogical reasoning \parencite{yang2024buffer,yu2023thought}, self-improvement \parencite{madaan2023self,tian2024toward}, abstraction \parencite{hong2024abstraction,zheng2023take}, search-based trial-and-error \parencite{besta2024graph,yao2023tree}, inference scaling \parencite{snell2024scaling,yang2025reasonflux}, and programming \parencite{gao2023pal,chen2022program}. Some recent works, such as Meta-COT \parencite{xiang2025towards} and REMA \parencite{wan2025rema}, explore enabling LLMs to acquire advanced thinking abilities such as meta-thinking \parencite{flavell1979metacognition}, which emphasizes strategic search and self-reflection. In this work, we propose a novel bilevel framework with the aim of stimulating the abstract modeling ability of the LLMs, enabling deeper insight, robust generalization, and clear reasoning processes.


\paragraph{Code-Enhanced LLM Reasoning}  
Code’s computational and logical strengths have made it a powerful approach for enhancing LLM reasoning. While some approaches leverage pseudocode to structure planning \parencite{wen2025codeplan}, reasoning logic \parencite{li2024chain}, or context tracking \parencite{puerto2024code}, their non-executable nature restricts access to the core powerful capabilities of code in terms of computation and tool invocation \parencite{wang2024executable}. Other works generate executable code through fine-tuning \parencite{gou2024tora,lu2024mathcoder2} or prompting \parencite{chen2022program,gao2023pal,xiao2024chain} for interpreter-based answering, yet they remain confined to narrow domains (e.g., math, symbolic tasks) due to the challenges in disentangling implicit reasoning signals from noisy, ambiguous information for abstract, complex input problems. Besides, some works \parencite{pan2023logic,ye2023satlm,zhou2024don} combine LLMs with logic programming (LP) languages, but these are restricted to logical reasoning tasks, and the unique syntax and interaction protocols of each LP environment severely limit their generality. Essentially, code serves as a form of executable symbolic logic, and our bilevel framework unleashes the power of code-augmented reasoning across a broader range of tasks.





%
\paragraph{Reinforcement Learning for LLMs}
To improve the performance and reliability, RLHF methods like Proximal Policy Optimization (PPO)\parencite{schulman2017proximal,ouyang2022training} are introducted for LLM alignment. While effective, PPO's reliance on a high-quality reward model and significant computational resources renders the alignment process complex and resource-intensive. Direct Preference Optimization (DPO)\parencite{rafailov2023direct} was then proposed to simplify this by directly optimizing with pairwise data, while its dependence on preference datasets limits its applicability and raises the risk of overfitting.Recent techniques like Group Relative Policy Optimization (GRPO)\parencite{shao2024deepseekmath} and REINFORCE++\parencite{hu2025reinforce++} aim to reduce the computational cost of critic networks in PPO, using group-based and batch-normalized baselines respectively. However, long contexts and extended inference trajectories remain impractical for smaller models, and to maintain flexibility, we often preferable to employ multiple models in a collaborative setup. Recently, bilevel optimization has emerged due to its popularity in various machine learning applications\parencite{liu2025argus,liu2025nested,chen2024robust}. In this paper, based on the principles of GRPO, we design a bilevel reinforcement learning optimization algorithm to jointly optimize our framework, achieving dynamic multi‑model collaboration with minimal overhead.

\section{Methodology}
\label{headings}

\subsection{Formalization}
We consider the general problem of natural language reasoning, where the input consists of  a textual problem description $q$ and a task instruction $o$. Given a large language model parameterized by  $\theta$, through step-by-step chain-of-thought(COT) process\parencite{wei2022chain}, the goal is to generate a coherent reasoning chain $r$  and a extracted final answer $a$. And through program-aided approaches  like program-of-thought(POT) \parencite{chen2022program} or PAL\parencite{gao2023pal}, the goal is to express reasoning steps into a  program $p$ and subsequently, this program $p$ is executed via a code interpreter to obtain the final answer $a$.
These processes can be expressed as the mapping:
\begin{equation}
\begin{aligned}
&\text{COT}: & (q, o) &\xrightarrow{\text{reasoning steps}} r \xrightarrow {\text{extract}}a \\
&\text{POT / PAL}: & (q, o) &\xrightarrow{\text{generate program}} p \xrightarrow{\text{code interpreter}} a
\end{aligned}
\end{equation}
However, these existing reasoning paradigms rely heavily on surface-level pattern recognition and lack structural modeling, which limitation has fueled critiques of LLMs as "stochastic parrots"\parencite{stechly2023gpt}. In contrast, humans rarely solve complex reasoning problems by directly decomposing them into step-by-step inference. Instead, human reasoning typically begins with modeling—such as the process of identifying variables, constraints, and objectives to represent a problem formally in operations research area. This modeling philosophy underpins the foundations of disciplines such as mathematics\parencite{bender2000introduction}, physics\parencite{newton1990mathematical}, operations research\parencite{churchman1957introduction}, and software engineering\parencite{brooks1995mythical}. It allows humans to grasp the underlying principles beyond surface patterns or memorization, enabling robust, efficient and generalizable reasoning capabilities.

%

Inspired by this cognitive pattern, we propose the \textit{Lang2Logic}  framework. Specifically, Lang2Logic decomposes reasoning into two steps specified as follows. 
\begin{itemize}
    \item \textbf{Optimization-Guided Formalization}: This step constructs a structured formal model $m$ from from the input $(q,o)$. Please note that the optimization-guided formalization framework aims to translate unstructured or semi-structured inputs—specifically a question $q$ and contextual observation $o$ into a structured model $m$. This model captures the underlying task logic and is designed to be interpretable, modular, and usable for downstream symbolic reasoning or computation. 
    \item \textbf{Logic Generation}: This step generates logical representations $p_l$ based on $m$ such as algorithmic logic or rule-based workflows, and then generate an executable symbolic workflow $p$ to compute the final answer $a$ from $p$. Code possesses strong logical representation capabilities. In this paper, we primarily use code as a typical executable symbolic workflow and employing the general-purpose programming language Python as the solver.
\end{itemize}
Unlike prior approaches that rely on direct code implementation or specific solvers\parencite{pan2023logic,ye2023satlm}, Lang2Logic enables models to understand problems at a abstract logical level, which outputs can be transcribed into any programming environment, enabling broader generalization and scalability. This process can be expressed as the mapping:
\begin{equation}
    \quad (q, o) \xrightarrow{\text{Formalization}} m  \xrightarrow{\text{Generation}} (p,a)
\end{equation}
We can formulate the generation of $p$ and $a$ using following joint probability:
\begin{equation}
P(m,p, a \mid q, o) = 
\underbrace{
\prod_{t=1}^{T_m} P(m_t \mid m_{<t}, q, o)
}_{\text{Modeling Phase}} 
\times 
\underbrace{
\prod_{t=1}^{T_p} P(p_t \mid p_{<t}, m)
\times
\prod_{t=1}^{T_a} P(a_t \mid a_{<t}, m,p)
}_{\text{Solving Phase}}
\end{equation}
where $T_m$, $T_p$, $T_a$ denote the sequence lengths (in tokens), $m_t$, $p_t$, $a_t$ denote token at position $t$
in the model sequence, solving logic sequence and executable program sequence,respectively and $m_{<t}$, $ p_{<t}$, $a_{<t}$ represent previous tokens. Here, final answer $a$ indicates program-generated outputs from external interpreter tool and $p$ indicates executable logic workflow.
\subsection{Overall Framework}
To effectively implement the \textit{Lang2Logic} paradigm, we partition reasoning into two specialized components—an \textit{Optimization-Guided Formalization LLM} for task abstraction and a \textit{Logic Generation LLM} for logic generation and execution. This design serves three primary purposes:(1) Cognitive Alignment — reflecting the human workflow (abstraction followed by computation); (2) Task Specialization — isolating the challenges of model construction from program generation, allowing each component to focus on its unique function and be optimized independently; (3) Context Efficiency — reducing the context burden on each component, making the framework more compatible with smaller models.
\paragraph{Optimization Oriented Problem Formalization} 
Inspired by principles from operations research, our approach adopts an \textbf{optimization perspective} to formalize reasoning problems. Specifically, the OGF LLM extracts the deep structural features of the reasoning task, identifying its core components such as \textit{variables, constraints, and objectives}. This process transforms complex, ambiguous natural language descriptions into structured, formal models, which are more amenable to precise reasoning and computation. As shown in figure \ref{fig:example}, this optimization-form modeling is versatile across diverse scenarios with several
advantages:\textit{(1)Reducing Ambiguity and Redundancy}: By filtering out irrelevant linguistic noise and capturing the essential problem components, the OGF mitigates the impact of ambiguity and redundancy in problem expressions and enhances the robustness of reasoning.\textit{(2)Abstracting Implicit Knowledge}: In many reasoning tasks, critical elements like objectives and constraints are not explicitly stated but must be inferred. The OGF is designed to identify these latent knowledge.For example, it should achieve the identification of the essence of problems such as causal inference, logical representation, probabilistic calculation, Boolean satisfiability problem (SAT), and constraint satisfaction problem (CSP) behind reasoning problems, and analyze the missing solution objectives and constraints in the input description, thus stimulating large language models to learn and apply abstract reasoning rules. 
\textit{(3)Enabling Clear and Interpretable Reasoning}:By formalizing the problem into a compact, structured representation, the OGF reduces the influence of overthing, context loss and hallucinations in Long COT,resulting in clearer, more interpretable reasoning process.

Formally, given the OGF LLM $\pi_{ogf}$, it takes the natural-language input $(q,o)$, and emits each reasoning problem into a structured model with five-tuple:
\begin{equation}
    m=(p,t,V,C,O)
\end{equation}
where \(p\) ("problem overview") is a brief summary of the task; \(t\) ("model type") indicates the model category (e.g., probabilistic calculate, SAT, CSP); \(V = \{v_1, \ldots, v_n\}\) is the set of decision or state variables; \(C = \{c_1, \ldots, c_k\}\) is the set of symbolic constraints (e.g., equations, inequalities, logical relations) over \(V\); and \(O\) ("objective") specifies the goal(s), such as “calculate \(x\)” or “determine if \(x\)=true”. By enforcing a canonical schema for $(p,t,V,C,O)$, the OGF LLM produces interpretable abstractions that capture core principles of the problem.
\paragraph{Logic Generation}
The LG llm $\pi_{lg}$ takes the structured model $m$ as input and generates the required solution logic $p_l$ and executable logic workflow $p$, executes it via a compiler, then returns the final answer $a$. 

It can use general-purpose programming languages (e.g. Python) and libraries (e.g., sympy, z3) to solve the models. In this paper, we utilize python as the executable symbolic representations to provide a unified action space for reasoning, computation, and external tool invocation, rather than relying on alternative output formats like JSON\parencite{wang2024executable}.
\paragraph{Collaborative Framework} To realize seamless cooperation between modeling and execution, we finally formulate the \textbf{Lang2Logic} framework as follows:
\begin{equation}
    \mathbf{a} \sim \pi_{\theta_{x}}(m|q) \cdot \pi_{\theta_{y}}(p_l,p,a|m )
    \label{eq:mot-formulation}
\end{equation}
where $\pi_{\theta_x}$ represents the OGF llm parameterized by $\theta_{x} \in \mathbb{R}^n$ and $\pi_{\theta_y}$ represents the LG llm parameterized by $\theta_{y} \in \mathbb{R}^n$. For any probem $q$ sampled from the set $Q$, we have $m \sim \pi_{\theta_{x}}(m|q)$ and $a\sim \pi_{\theta_{y}}(p_l,p,a|m)$.

Decoupling abstraction from execution allows our collaborative framework to balance flexibility and robustness in reasoning. The interaction between the OGF and the LG is inherently bidirectional, supporting iterative refinement. If the LG encounters errors or unsatisfactory outputs, it can determine either regenerating the program or giving feedback to the OGF to adjust the model. This mechanism ensures a crucial aspect of complex problem-solving:
the dynamic interplay between planning and execution.

\subsection{Model-augmented Supervised Fine-tuning}
\paragraph{Constructing Model-Augmented Dataset}
To better align the OGF with our reasoning format, we collect a "model-augumented" dataset to fine-tune the OGF LLM for cold start. We use code-enhanced rejection sampling for obtaining a high-quality dataset $D_{mod}$\parencite{guan2025rstar}. In short,for each query $q$ sampled from seed corpus dataset $D$,we apply a best-of-$N$ sampling strategy: prompt an SOTA larger LLM to generate $N$ candidate triples $\langle \text{think} \rangle, \langle \text{model} \rangle, \langle \text{code} \rangle$. Then, we extract the $\langle \text{code} \rangle$ blocks and executing them in a sandboxed interpreter, and only those whose outputs match the ground truth $a$ are kept. Since we do not prioritize reasoning path diversity in our dataset construction, so then we propose a "preference self-evaluation" methods: prompt the SOTA LLM to evaluate the remaining valid samples on clarity and conciseness, and we retain the top-K exemplars per problem. More details can be find in Appendix A.

This pipeline yields a high-quality, model-augmented training dataset $D_{mod}$ for fine-tuning the modeling LLM, which can be defined as:
\begin{equation}
    D_{mod}=\{q,a,s,m,p|(q,a) \in D,\text{Exec}(p)=a,\text{Rank}(m)\}
\end{equation}
where $q$ is the input problem and $a$ is the ground truth sampled from the original dataset $D$. $s$, $m$ and $p$ respectively represents the analysis steps of modeling, the final structured model abstracted from $S$ and executed programs.The construction of \(D_{\mathrm{mod}}\) enforces two critical constraints:  \(\text{Exec}(P) = a\), the execution of program must equal the ground-truth answer $a$; \(\text{Rank}(M) \), retaining only the top-$\min(K, I)$ ranked $M$, where K represents the pre-defined number of samples to be retained, and I represents the number of samples with correct results.

\paragraph{Model-augmented Supervised Fine-tuning}
Inspired by DeepSeek-R1\parencite{guo2025deepseek}, we first perform a cold-start fine-tuning of the Modeling LLM on our model-augmented dataset $D_{\mathrm{mod}}$ before after joint bilevel training. Concretely, given each sample $(q,m)\in D_{\mathrm{mod}}$, we train the Modeling LLM $\pi_{\theta_x}$ to maximize the likelihood of the target model tokens, which be formulated as:
\begin{equation}
\mathcal{L}_{\text{SFT}}(\theta_x)
= -\sum_{(q,s,m)\in D_{\mathrm{mod}}} \log \pi_{\theta_x} (s,m|q).
\end{equation}
This supervised fine-tuning step better aligns the modeling LLM with our reasoning paradigm and lays the foundation for subsequent joint reinforcement training.

\subsection{Bilevel Reinforcement Learning Algorithm}
\paragraph{Bilevel Optimization Problem Formulation}


Bilevel optimization~\parencite{jiao2023asynchronous,jiao2025pr} offers a powerful and flexible framework for modeling hierarchical decision-making processes, in which one optimization problem is nested within another. By capturing both adversarial and cooperative interactions, it is closely related to several important paradigms, including robust optimization~\parencite{yang2008distributed,yang2014distributed}, Stackelberg game\parencite{bruckner2011stackelberg}, and meta-learning\parencite{franceschi2018bilevel,jiao2023asynchronous}.  

Motivated by the flexibility and expressive power of this formulation, we adopt a bilevel perspective to model the interaction between \textit{task modeling} and \textit{problem solving}, while leveraging reinforcement learning to improve generalization to novel tasks. We design our bilevel reinforce learning algorithm based on \textit{Group Relative Policy Optimization(GRPO)}\parencite{shao2024deepseekmath}. As shown in Eq. \eqref{eq:mot-formulation}, our framework factorizes the reasoning process into high-level and low-level policies. 

We treat $\pi_{\theta_x}$ (the OGF LLM) as the high-level policy and $\pi_{\theta_y}$ (the LG LLM) as the lower-level policy. For each problem $q\sim P(Q)$,  the upper‐level policy first samples $G$ candidate models $\{m_i\}$ from $\pi_{\theta_x}^{old}(m|q)$, and the lower-level policy then samples $P$ candidate outputs $\{o_{ij}\}$ from $\pi_{\theta_y}^{old}(o|m_i)$. Considering the GRPO objective(without KL penalties), the upper-level(OGF) updates maximizes:
\begin{equation}
    \begin{split}
\mathcal{J}_h ({\theta_x}) &= \mathbb{E}_{q \sim P(Q), \{m_i\}_{i=1}^G \sim \pi_{\theta_x^{\text{old}}}(\cdot|q)} 
\\ &\Bigg[ \frac{1}{G} \sum_{i=1}^G \Bigg(
 \min \left( \frac{\pi_{\theta_x}(m_i|q)}{\pi_{\theta_x^{\text{old}}}(m_i|q)} A_{m_i}, \text{clip} \left( \frac{\pi_{\theta_x}(m_i|q)}{\pi_{\theta_x^{\text{old}}}(m_i|q)}, 1-\epsilon, 1+\epsilon \right) A_{m_i} \right)  \Bigg) \Bigg]
\end{split}
\label{eq:modeling-grpo}
\end{equation}
where $\epsilon$ is clip hyper-parameters and $A_{m_i}$ is the advantage of upper-level.
The lower-level (LG) update similarly maximizes:
\begin{equation}
    \begin{split}
\mathcal{J}_l(\theta_y) &= \mathbb{E}_{q \sim P(Q), m_i \sim \pi_{\theta_x}(\cdot|q), \{o_{ij}\}_{j=1}^P \sim \pi_{\theta_y^{\text{old}}}(\cdot|m_i)}  \\
& \Bigg[\frac{1}{P} \sum_{j=1}^P \left( \min \left( \frac{\pi_{\theta_y}(o_{ij}|m_i)}{\pi_{\theta_y^{\text{old}}}(o_{ij}|m_i)} A_{o_{ij}}, \text{clip} \left( \frac{\pi_{\theta_y}(o_{ij}|m_i)}{\pi_{\theta_y^{\text{old}}}(o_{ij}|m_i)}, 1-\epsilon, 1+\epsilon \right) A_{o_{ij}} \right) \right) \Bigg]
\end{split}
\label{eq:coding-grpo}
\end{equation}
where $A_{o_{ij}}$ is the advantage of lower-level. Together, the bilevel optimization objectives for our framework can be formulated as:
\begin{equation}
\begin{aligned}
\max_{\theta_x}& \quad \mathcal{J}_h(\theta_x, \theta_y^*) \\
\text{s.t. }&\quad \theta_y^*=\arg\max_{\theta_y} \mathcal{J}_l(\theta_x,\theta_y )
\end{aligned}
\end{equation}
\paragraph{Rule-based Reward Modeling}
Following the spirit of R1\parencite{guo2025deepseek}, we also employ a rule-based reward that combines accuracy—correctness of the answer and format—enforces the model to put its output in given format.The advantage functions of the upper and lower models adopt normalized rewards. Specifically, each lower-level sample $o_{ij}$ receives a reward $r_o(o_ij)$ reflecting both answer correctness and format correctness. The lower level LLM's advantage $A_{o_{ij}}$ is:
\begin{equation}
    A_{o_{ij}} = \frac{r_o(o_{ij}) - \text{mean}_{k=1}^P r_o(o_{ik})}{\text{std}_{k=1}^P r_o(o_{ik})}
\end{equation}
For OGF LLM, the reward $r_m(m_i)$ is also defined as the mean of its P lower-level rewards, and its advantage is:
\begin{equation}
    A_{m_i} = \frac{r_m(m_i) - \text{mean}_{k=1}^G r_m(m_k)}{\text{std}_{k=1}^G r_m(m_k)}
\end{equation}
Specifically, in the reward function $r_m(m_i)$ for the upper-level model, its correctness reward is taken as the average of the rewards of the results sampled from the lower-level model.

\paragraph{Bilevel Reinforcement Learning Algorithm}
We solve the bilevel optimization problem via an alternating update methods\parencite{schulman2017proximal,jian2024tri}: first update $\pi_{\theta_y}$, holding $\theta_x$ fixed, then update $\pi_{\theta_x}$ using the newly updated $\theta_{y}$, as summarized in Algorithm~\ref{alg:bilevel-grpo}

\begin{algorithm}[t]
\caption{Alternating Bilevel RL for Lang2Logic}
\label{alg:bilevel-grpo}
\KwIn{Policies $\pi_{\theta_x},\pi_{\theta_y}$; data $D$; clip $\epsilon$; batches $B,G,P$; loops $N_{\ell},N_{h}$}
\KwOut{Optimized $\theta_x,\theta_y$}
\BlankLine
\For{$\!iter=1\dots I\!$}{
  \tcp{-- LG update --}
  \For{$k=1\dots N_{\ell}$}{
    Sample $\{q_i\}_{i=1}^B\!\sim\!D$, then for each $q_i$:
    \(\,m_i\sim\pi_{\theta_x^{\rm old}}(m|q_i)\), 
    \(\{o_{ij}\}_{j=1}^P\sim\pi_{\theta_y^{\rm old}}(o|m_i)\);
    compute $r_o(o_{ij}),A_{o_{ij}}$  
    \vspace{1pt}\\
    Update $\theta_y$ via Eq. \eqref{eq:coding-grpo} using $A_{o_{ij}}$
  }
  \tcp{-- OGF update --}
  \For{$k=1\dots N_{h}$}{
    Sample $\{q_i\}_{i=1}^B\!\sim\!D$, then for each $q_i$:
    \(\{m_{i}\}_{i=1}^G\sim\pi_{\theta_x^{\rm old}}(m|q_i)\); 
    for each $m_{i}$ sample $\{o_{ij}\}\sim\pi_{\theta_y}(o|m_{i})$, 
    compute $r_m(m_{i}),A_{m_{i}}$  
    \vspace{1pt}\\
    Update $\theta_x$ via Eq. \eqref{eq:modeling-grpo} using $A_{m_i}$
  }
}
\end{algorithm}

\section{Experiments}
\paragraph{Base Models,Training Data and Experimental Setup}
We implement MoT on two model scales: qwen2.5-7B-instruct\parencite{yang2024qwen2} and qwen2.5-1.5B-instruct\parencite{yang2024qwen2}. In SFT stage for modeling LLM, we select subtasks from Flan-Collection\parencite{longpre2023flan}, including gsm8k\parencite{cobbe2021training},aqua\parencite{ling2017program},qed\parencite{lamm2021qed} as source dataset, and prompt qwen-turbo\parencite{yang2024qwen2} with reject sampling and self-evaluation to collect $D_{mod}$ with 28K samples.We conduct initialization training for 2 epochs
using an AdamW optimizer along with the cosine learning rate scheduler. To further strengthen the collaboration between two models, we conduct our bilevel optimization algorithm on gsm8k\parencite{cobbe2021training} for 5 iterations with unsloth framework for efficient training, similar using an AdamW optimizer along with the cosine learning rate. We using NVIDIA GeForce RTX 4090 or A100 for training, depending on models. More details in Appendix B.

\paragraph{Evaluation Datasets}
We evaluate MoT across a diverse range of critical reasoning tasks and datasets without any training. We mainly focus on those "hard reasoning" tasks, which means the model cannot direct obtain the answer by prior knowledge or common sense: (1) \textbf{Causal reasoning}: (a) Cladder\parencite{jin2023cladder}, a comprehensive dataset that covers associational, interventional and counterfactual three types casual reasoning problems. (b)CounterBench\parencite{chen2025counterbench}, which requires multi-step inference based on given information. (2)\textbf{Logical reasoning}: (a)K\&K puzzles\parencite{xie2024memorization}, a logic puzzle independent of commonsense interference with different difficulty levels. (b)Autologi\parencite{zhu2025autologi}, an open-ended logical reasoning dataset, unaffected by random conjecture. (3)\textbf{Mathematical Reasoning}: (a)GSM-hard\parencite{gao2023pal},a harder version of GSM8K\parencite{cobbe2021training}with larger numbers that are less common. (b)SVAMP\parencite{patel2021nlp}, a robustness evaluation benchmark. (4)\textbf{Spatial Reasoning}: (a)Next Step Prediction, a subtask of visual navigation\parencite{wu2024mind} requiring multi-hop spatial reasoning, while the former is more complex. (5)\textbf{Temporal Reasoning}: (a) ToT-arithmetic, a subtask of Test-of-Time benchmark\parencite{fatemi2024test} requiring time arithmetic operations under a story of real-world. (b)TimeQA\parencite{chen2021dataset}, a comprehensive time-sensitive QA dataset that requires time understanding and logical reasoning.

\paragraph{Baselines and Evaluations}
We consider the following baselines, which respectively represent different reasoning paradigms: (1)Standard CoT\parencite{wei2022chain}: elicits LLM to analyze natural language problems and generate intermediate reasoning steps. (2)Plan-and-Solve\parencite{wang2023plan}: A two‐stage approach that first generates a high‐level plan and then generate the response following the plan.(3)Self‐Refine\parencite{madaan2023self}:An iterative reasoning strategy that alternates between generating answers and refining them based on self-evaluation feedback.(4)PAL\parencite{gao2023pal}: uses the LLM to read natural language problems and generate programs as the intermediate reasoning steps. To ensure answer formats are easily verifiable, we evaluate all methods in a few‐shot setting (1–3 shots) without any training on the evaluation benchmarks, and apply identical configurations to all baselines. Additional results using in‐domain training data are provided in the Appendix C.

\subsection{Main Results}
As highlighted in Table 1 and Table 2. We verified the breakthrough performance of the Lang2Logic framework in complex reasoning tasks through multi-dimensional benchmark tests. The framework demonstrates two core advantages over traditional approaches:
\vspace{-8pt}
\paragraph{Cross-domain reasoning} Lang2Logic significantly outperformed CoT (55.8\%) and PAL (55.4\%) with 63\% accuracy on the AUTOLOGI problem, a 12.9\% improvement over the suboptimal baseline. With an accuracy of 83.5\% in the temporal test dataset, it was 11.3\% higher than PAL (75\%) and 35\% higher than CoT (61\%). A 48.3\% accuracy was achieved on the geometric prediction problem, compared to 41.1\% for PoT and 37.4\% for planning and solving, an improvement of 17.5\% relative to the recent baseline.
\vspace{-8pt}
\paragraph{Enhanced performance on complex reasoning problems} Lang2Logic also showed differential performance on mathematical reasoning benchmarks. An accuracy of 82\% was achieved on the complex GSM dataset, an improvement of 14.4 percentage points over the sub-optimal benchmark, while an accuracy of 92.3\% was achieved on the simple SVAMP, and the accuracy of 14.4 percentage points on the complex GSM dataset was better than the sub-optimal benchmark, a slight improvement of 3.7 percentage points over the sub-optimal benchmark. This performance difference highlights the adaptive nature of the framework in dealing with the complexity of different problems. The significant improvement in GSM highlights the effectiveness of Lang2Logic in solving challenging mathematical reasoning tasks that require multistep logical operations, and highlights the potential of Lang2Logic to improve the performance of mathematical reasoning tasks, the limited progress on SVAMP is consistent with the lower complexity inherent in datasets and the use of explicit logical rules for structured problem representations with rich properties.

In short, the Lang2Logic framework breaks through the three dimensions of interpretability, cognitive efficiency, and task adaptability through an innovative two-stage reasoning paradigm. The experimental data show that it shows significant advantages in complex scenarios such as combinational logic, spatio-temporal reasoning, and mathematical problems, especially in difficult tasks that require multi-step reasoning. This structured reasoning mechanism not only breaks through the performance bottleneck of the traditional single-stage method, but also provides a new path for the verification and optimization of the subsequent reasoning process. We further analyzed the experimental results and error patterns. For more figures and tables, please refer to Appendix D.

\begin{table}[ht]
\centering
\caption{Performance of different models across Mathematical and Temporal tasks with improvements from Lang2Logic. Among them, GSM8K is an in-domain dataset, and the rest are out-of-domain datasets.}
\label{tab:model_performance}
\begin{tabular}{lcccccc} 
\toprule
\multirow{2}{*}{\textbf{Model}} & \multicolumn{3}{c}{\textbf{Mathematical Reasoning}} & \multicolumn{2}{c}{\textbf{Temporal Reasoning}} \\
\cmidrule(lr){2-4} \cmidrule(lr){5-6}  
 & \cellcolor{myblue}GSM8K & GSM\_Hard & SVAMP & Test-of-Time & TimeQA \\
\midrule
\textbf{qwen2.5-7B} & 91.6 & 59.3 & 89.5 & 61 & 45.6  \\
Plan-and-Solve & 90.1 & 58 & 87.7 & 41.31 & 38.8 \\
Self-Refine & 90.9 & 57.3 & 88.3 & 60.5 & 44.0 \\
PAL & \underline{92.4} & \underline{71.7} & \underline{90.0} & \underline{75} & \underline{50.5}   \\
Lang2Logic(ours) & \textbf{94.3} & \textbf{82} & \textbf{93.3} & \textbf{83.5}& \textbf{59.4}  \\
(Relative $\Delta$ Gain) & \textcolor{red}{(+2.9\%)} & \textcolor{red}{(+14.4\%)} & \textcolor{red}{(+3.7\%)} & \textcolor{red}{(+11.3\%)} & \textcolor{red}{(+17.6\%)} \\
\midrule
\textbf{qwen2.5-1.5B} & 73.2 & 29.3 & \underline{78.3} & 15.0 & \underline{16.6} \\
Plan-and-Solve & \underline{74.4} & 25.7 & 74.7 & 12.5 & 14.0  \\
Self-Refine & 69.8 & 29.5 & 72.3 & 20.3 & 14.0  \\
PAL & 72.6 & \underline{51.7} & 73.7 & \underline{31} & 16.3  \\
Lang2Logic(ours) & \textbf{78.5} & \textbf{56} & \textbf{80.7} & \textbf{39.5}& \textbf{21.5}  \\
(Relative $\Delta$ Gain) & \textcolor{red}{(+5.5\%)} & \textcolor{red}{(+8.3\%)} & \textcolor{red}{(+3.1\%)} & \textcolor{red}{(+27.4\%)} & \textcolor{red}{(+28.8\%)}  \\
\midrule
\bottomrule
\end{tabular}
\end{table}

\begin{table}[ht]
\centering
\caption{Performance of different models across Causal,Logical and Spatial tasks with improvements from Lang2Logic.}
\label{tab:model_performance_2}
\begin{tabular}{lcccccc}
\toprule
\multirow{2}{*}{\textbf{Model}} & \multicolumn{2}{c}{\textbf{Causal reasoning}} & \multicolumn{2}{c}{\textbf{Logical Reasoning}}& \multicolumn{1}{c}{\textbf{Spatial Reasoning}} \\
\cmidrule(lr){2-3} \cmidrule(lr){4-5} \cmidrule(lr){6-6}
 & Cladder & CounterBench  & AutoLogi & K\&K puzzles & Next Step Prediction\\
\midrule
\textbf{qwen2.5-7B} & 56.4 & 41.7 & \underline{55.8} & 23.1 & 37\\
Plan-and-Solve & 54,4 & 47.3 & 54.1 & 14.9 & 37.4 \\
Self-Refine & 55.0 & 50.1 & 51.5 & 22.7 & 35.7\\
PAL & \underline{60.0} & \underline{65.2} & 55.4 & \underline{27.1} &\underline{41.1}\\
Lang2Logic(ours) & \textbf{69.0} & \textbf{82.5} & \textbf{63} & \textbf{36.7}& \textbf{48.3}\\
(Relative $\Delta$ Gain) & \textcolor{red}{(+15.0\%)} & \textcolor{red}{(+23.5\%)} & \textcolor{red}{(+12.9\%)} & \textcolor{red}{(+35.4\%)} &\textcolor{red}{(+17.5\%)}\\
\midrule
\textbf{qwen2.5-1.5B} & 44.0 & 49.0 & \underline{28.7} & \underline{5.8} & 22.1\\
Plan-and-Solve & 45.1 & \underline{49.5} & 15.1 & 4.3 & 24.6  \\
Self-Refine & 42.0 & 46.7 & 25.1 & 5.2 & 27.9 \\
PAL & \underline{52.8} & 48.3 & 25.7 & 5.6 & \underline{31.6}\\
Lang2Logic(ours) & \textbf{58.0} & \textbf{62.0} & \textbf{29.4} & \textbf{8.1}& \textbf{35.1} \\
(Relative $\Delta$ Gain) & \textcolor{red}{(+9.8\%)} & \textcolor{red}{(+25.3\%)} & \textcolor{red}{(+2.4\%)} & \textcolor{red}{(+39.7\%)} &\textcolor{red}{(+11.1\%)} \\
\midrule
\bottomrule
\end{tabular}
\end{table}

\section{Conclusion}
In this paper, we proposed \textit{Lang2Logic}, a bilevel structured reasoning framework that aims to bridge the gap between natural language understanding and logical formalization. Our approach explicitly separates task abstraction from logic generation, allowing LLMs to construct structured formal models and then generate executable symbolic workflows. This design captures the underlying logic of complex problems more effectively, facilitating more accurate, interpretable, and generalizable reasoning. We introduced a two-stage training strategy, including a supervised cold-start phase using a model-augmented dataset and a bilevel reinforcement learning algorithm for joint optimization. This training regimen enables the framework to effectively coordinate abstraction and execution, significantly enhancing its reasoning performance. Extensive experiments across diverse reasoning benchmarks, demonstrated that our framework consistently outperforms different baselines, achieving over 10\% average accuracy gains. These results confirm that our Lang2Logic can offer a promising direction for the future of systematic reasoning with LLMs.

\printbibliography
\end{document}